\documentclass[twoside]{article}
\usepackage[accepted]{aistats2016}

\usepackage{bm}
\usepackage{hyperref}
\usepackage{url}
\usepackage{amsmath,amssymb,amsfonts,amsthm,xfrac}
\usepackage{capt-of}
\usepackage{stfloats}
\usepackage{subfigure} 
\usepackage{wrapfig}
\usepackage{natbib}
\usepackage{todonotes}
\newcommand{\norm}[1]{\lVert#1\rVert}

\newcommand{\iid}{\overset{iid}{\sim }}

\newcommand{\ud}{\mathrm{d}}
\newcommand{\mi}{\mathrm{i}\,}

\newcommand{\bc}{\beta^{\mathrm{cos}}}
\newcommand{\bs}{\beta^{\mathrm{sin}}}
\newcommand{\bmap}{\beta_{\mathrm{MAP}}}

\newcommand{\calD}{\mathcal{D}}

\newcommand{\calN}{\mathcal{N}}
\newcommand{\calX}{\mathcal{X}}
\newcommand{\bK}{\mathbf{K}}
\newcommand{\E}{\mathbb{E}}

\newcommand{\R}{\mathbb{R}}
\newcommand{\W}{W}

\newcommand{\where}{\mathrm{where }}

\newcommand{\tand}{\mathrm{ and }}

\newtheorem{theorem}{Theorem}[section]

\newtheorem{proposition}[theorem]{Proposition}
\theoremstyle{definition}

\theoremstyle{remark}

\usepackage{lipsum}

\newcommand\blfootnote[1]{%
  \begingroup
  \renewcommand\thefootnote{}\footnote{#1}%
  \addtocounter{footnote}{-1}%
  \endgroup
}

\usepackage{etoolbox}
\patchcmd{\maketitle}{}{}{}{}

\begin{document}

\twocolumn[

\aistatstitle{Bayesian Nonparametric Kernel-Learning}

\aistatsauthor{ Junier B. Oliva$^*$ \And Avinava Dubey$^*$ \And Andrew G. Wilson \And Barnab{\'a}s P{\'o}czos  \AND  Jeff Schneider \And Eric P. Xing }

\aistatsaddress{Machine Learning Department \\
Carnegie Mellon University\\
Pittsburgh, PA 15213 \\
\texttt{ \{joliva, akdubey, andrewgw, bapoczos, schneide, epxing\}@cs.cmu.edu} } ]


%

\begin{abstract}
\vspace{-0.05in}
Kernel methods are ubiquitous tools in machine learning. 
However, there is often little reason for the common practice of selecting a kernel a priori. Even if a universal approximating kernel is selected, the quality of the finite sample estimator may be greatly affected by the choice of kernel. Furthermore, when directly applying kernel methods, one typically needs to compute a $N \times N$ Gram matrix of pairwise kernel evaluations to work with a dataset of $N$ instances. The computation of this Gram matrix precludes the direct application of kernel methods on large datasets, and makes kernel learning especially difficult. 

In this paper we introduce Bayesian nonparmetric kernel-learning (BaNK), a generic, data-driven framework for scalable learning of kernels. 
BaNK places a nonparametric prior on the spectral distribution of random frequencies allowing it to both learn kernels and scale to large datasets.
We show that this framework can be used for large scale regression and classification tasks. Furthermore, we show that BaNK outperforms several other scalable approaches for kernel learning on a variety of real world datasets.\blfootnote{$*$ denotes equal contribution} 
\end{abstract}\vspace{-0.2in}

\section{Introduction}\vspace{-0.05in}
Kernel methods such as support vector machines (SVMs), kernel-ridge regression, kernel-PCA, and Gaussian processes (GPs) have become the cornerstone of many machine learning approaches.
However, the choice of kernel, which profoundly affects performance, has until recently received little attention. 
In fact, finite sample estimates will be affected by the kernel choice notwithstanding the use of an universal approximating kernel.
Indeed, the a priori choice of a fixed kernel in kernel methods is typically ad-hoc and not data-driven.
Even when learning kernel hyperparameters,
one is typically limited to an arbitrarily chosen and restrictive family
of kernel functions, exploring only a very small subset of reasonable
possibilities.
Given that the choice of kernel is an important free parameter in kernel methods, and generally there are few a priori reasons for kernel selections, a principled and data-driven method for learning kernels is extremely useful.


Furthermore, kernel methods often do not scale to datasets with a large number of instances due to the need to compute and store an $N \times N$ Gram matrix, $\bK$, for $N$ training points. Moreover, kernel methods, such as GPs, will often require manipulations of $\bK$ like solving linear systems and computing log determinants, leading to a $O(N^3)$ time complexity. Considering that modern datasets are only increasing in size, and complicated machine learning tasks require large datasets, it is vital to mitigate the high computational cost of kernel methods.

In order to provide a method that scales to large datasets and adaptively learns the kernel to use in a data-driven fashion, this paper presents the \textit{Bayesian nonparametric kernel-learning} (BaNK) framework. BaNK is a novel approach that will use random features to both provide a scalable solution and learn kernels. This approach scales through random features and places a Bayesian nonparametric distribution over kernels, with support for any stationary kernels.

\begin{figure}
\vspace{-0.275in}
\centering
    \subfigure[RKS]{{\includegraphics[width=0.2\textwidth]{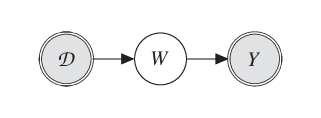}}}
    \subfigure[BaNK]{{\includegraphics[width=0.2\textwidth]{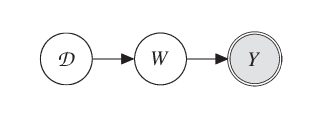}}}\vspace{-0.075in}
    \caption{(a) Traditional random feature approach where the distribution $\calD$ of random features $W$ is held fixed. (b) BaNK framework where $\calD$ is random. } \label{fig:cartoon} \vspace{-0.05in} 
\end{figure}

Random features have been recently shown to be an effective way to scale kernel methods to large datasets. Roughly speaking, random feature techniques like random kitchen sinks (RKS) \citep{rahimi2007random} work as follows. Given a shift invariant kernel $K(x,x^\prime)=k(x-x^\prime)$, one constructs an approximate primal space to estimate kernel evaluations $K(x,x^\prime)$ as the dot product of finite vectors $\varphi(x)^T\varphi(x^\prime)$. The vectors $\varphi$ are constructed with random frequencies drawn from a distribution $\calD$ that is defined by $K$. Similarly, a distribution $\calD$ from which random frequencies are drawn from defines a kernel $K$ that the random frequencies approximate. It is this last observation that is key for the BaNK framework. 
BaNK will allow the distribution $\calD$ to vary with the given data, effectively learning the kernel. In particular, BaNK shall vary $\calD$ with a graphical model approach where we treat $\calD$ as a latent parameter and place a prior on it (Figure \ref{fig:cartoon}). The prior on $\calD$, along with the data generation model, will allow one to sample from a posterior over $\calD$ in order to learn the corresponding kernel.
\vspace{-0.1cm}
We model $\calD$ as a mixture of Gaussians with a Dirichlet process prior, which allows BaNK to learn a kernel from a rich, broad class. Furthermore, with the use of random features, we are able to efficiently sample the model parameters and work over larger datasets. Moreover, by using Metropolis-Hastings we sample from a proper posterior, thus the kernels we learn are interpretable since the random features are asymptotically guaranteed to come from the underlying posterior distribution unlike greedy non-convex optimization methods.

\textbf{Outline} The rest of this paper is structured as follows. First we review the use of random features for kernel approximation and show how such an approach can be used for flexible and efficient kernel learning. Second, we detail our graphical model framework both for supervised regression and classification tasks. Third, we expound on our inference method for sampling from the model posterior. Forth, we illustrate the use and performance of BaNK for both regression and classification on several datasets. Lastly, we cover related works and give concluding remarks.

\vspace{-0.1in}\section{Model}\vspace{-0.05in}

\subsection{Random Features for Kernel Estimation}\vspace{-0.05in}
\label{sec:randfeat}
Below we briefly review the method of random Fourier features for the approximation of kernels \citep{rahimi2007random}. The details of the method will help motivate and explain our BaNK model. Henceforth, we will only consider continuous shift-invariant kernels defined over $\R^d$: $K(x,y)=k(x-y)$ where $x,y\in\R^d$ and $k$ is a positive definite function. The use of random Fourier features for kernel approximation is a result of Monte Carlo integration using Bochner's theorem \citep{rudin1990fourier}. Bochner's theorem states that a continuous shift-invariant kernel $K(x,y)=k(x-y)$ is a positive definite function if and only if $k(t)$ is the Fourier transform of a non-negative measure $\rho(\omega)$. Note further, that if $k(0)=1$, then $\rho(\omega)$ will be a normalized density. That is, if we define $\zeta_{\omega}(x) \equiv \exp(\mi \omega^T x)$, then
\begin{align}
k(x-y) &= \int_{\R^d} \rho(\omega) \exp\left(\mi \omega^T (x-y)\right) \ud \omega \nonumber \\
&= \E_{\omega\sim\rho}[\zeta_{\omega}(x) \zeta_{\omega}(y)^*].
\end{align}
Hence, using Monte Carlo integration, we can approximate $K(x,y) = k(x-y)$ using $\omega_j \iid \rho$:
\begin{align}
k(x-y) \approx \frac{1}{M} \sum_{j=1}^M \zeta_{\omega_j}(x) \zeta_{\omega_j}(y)^* \label{eq:bochners}.
\end{align}
In particular, if our kernel $k$ is real-valued, then we can discard the imaginary part of \eqref{eq:bochners}:
\begin{align}
& k(x-y) \approx \varphi(x)^T \varphi(y),\ \varphi(x) \equiv  \label{eq:rksfeats} \\ 
 & \tfrac{1}{\sqrt{M}}[\cos(\omega_1^T x),\ldots,\cos(\omega_M^T x),\sin(\omega_1^T x),\ldots,\sin(\omega_M^T x)]^T. \nonumber
\end{align}
The great advantage of such an approximation is that we may now estimate a function in the RKHS as a linear operator in the random features: $f(x) = \sum_{i=1}^m \alpha_i K(x_i,x) \approx \sum_{i=1}^m \alpha_i \varphi(x_i)^T \varphi(x) = \psi^T\varphi(x)$, where $\psi \equiv \sum_{i=1}^m \alpha_i \varphi(x_i)$. Thus we may work directly in a primal space of $\varphi(x)$ and avoid computing large Gram matrices. To recap, using the approximation of kernels with random features works as follows: choose a kernel defined by $k$ (with $k(0)=1$), take its Fourier transform, $p(\omega)$, which will be a pdf over $\R^d$; draw $M$ i.i.d. samples from $\rho(\omega)$, $\{\omega_j\}_{j=1}^M$; estimate the kernel with $K(x,y) \approx \varphi(x)^T \varphi(y)$ as in \eqref{eq:rksfeats}.


However, Bochner's theorem also allows one to work in the other direction. That is, we may start with a distribution $\calD$ with pdf $\rho(\omega)$ and take the characteristic function (the inverse Fourier transformation) to define a shift-invariant kernel $k$. For example, suppose that $\rho(\omega) = \calN(\omega | \mu, \Sigma)$, where $\calN(\omega | \mu, \Sigma)$ is the pdf of $\calN(\mu,\Sigma)$. Taking its characteristic function we see that $k(t) = \exp\left( \mi \mu^Tt - \tfrac{1}{2} t^T \Sigma t\right)$ would be the corresponding shift-invariant kernel. From the kernel learning perspective, Bochner's theorem yields an object to manipulate for the learning of one's kernel: $\rho(\omega)$ the distribution of random features. 

We consider distributions that are mixtures of Gaussians:
\vspace{-0.1cm}
{\small
\begin{align}
&\rho(\omega) = \sum_{\ell=1}^L \pi_\ell \calN(\omega | \mu_\ell, \Sigma_\ell) \rightarrow  \nonumber \\
&k(t) = \sum_{\ell=1}^L \pi_\ell \exp\left( \mi \mu_\ell^T t - \tfrac{1}{2} t^T \Sigma_\ell t\right) \label{eq:mixkern}.
\end{align}
}
This makes for very general kernels; for a discussion on general properties of these kernels for finite $L$ please see \citep{Wilson2013}. In fact, i) noting that Gaussian mixture models are universal approximators of densities and may hence approximate any spectral distribution, and ii) using Plancherel's Theorem to relate spectral accuracies to the original domain \citep{silverman1986density, yang2015carte} it follows that:
\begin{proposition}
The expression of $\rho(\omega)$ in \eqref{eq:mixkern} can approximate any shift invariant kernel. 
\end{proposition}

For our applications we only need real-valued kernels,
hence we use the real part of \eqref{eq:mixkern}:
\begin{align}
K(x,y) =& \sum_{\ell=1}^L \pi_\ell \exp\left(- \tfrac{1}{2} (x-y)^T \Sigma_\ell (x-y)\right) \nonumber \\ 
&\quad\qquad \cos\left( \mu_\ell^T (x-y) \right) \\
\approx& \varphi(x)^T \varphi(y),
\label{eq:approxmix}
\end{align}
where $\varphi(x)$ is as in \eqref{eq:rksfeats}.
An application of the random feature approximation bounds found in \citep{rahimi2007random,le2013fastfood,sutherl2015error} yields that:
\begin{proposition}
For compact $\calX \subset \R^d$ with finite diameter, we have that
\begin{align*}
&\Pr\left[\sup_{x,y\in\calX} |K(x,y)- \varphi(x)^T\varphi(y)| \geq \epsilon \right] \\
&= O\left( \frac{1}{\epsilon^2}\exp\left( \frac{-M\epsilon^2}{4(d+2)} \right)  \right).
\end{align*}
\end{proposition}

Using the above, it may be seen that one can effectively approximate shift-invariant kernels using random features drawn from Gaussian mixtures. However, in order to learn the kernel, one still needs a mechanism to determine the Gaussian mixture to use. We take a graphical model approach to determine the mixture for $\rho(\omega)$ in a principled, data-driven fashion. 

\vspace{-0.05in}
\subsection{Graphical Model}\vspace{-0.05in}
As described above, one may vary and tune kernels with the choice of density over random features, $\rho(\omega)$. Thus, in our model, we take this distribution itself to be a random latent parameter, in effect placing a prior over all stationary kernels, resulting in a strictly more general method than the traditional approach of using a fixed RBF kernel. 

Roughly speaking, the BaNK model will consist of three major parts: one, a prior for stochastically generating the random feature distribution $\rho(\omega)$; two, a prior for the generation of the parameters of a linear model in the primal space of random features; three, a generative data model with noise to generate labels given input covariates and the rest of the parameters. 

\begin{figure}[t]
    \centering
    \includegraphics[width=0.45\textwidth]{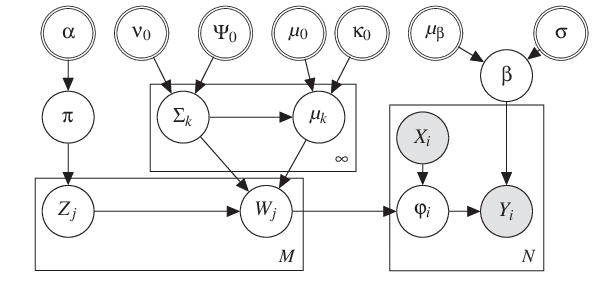}
    \caption{Plate diagram for the graphical model for BaNK learning framework.}
    \label{fig:awesome_image}
\end{figure}
\vspace{-0.3cm}

First, the spectral distribution $\rho(\omega)$ is generated. As previously mentioned, a robust and flexible choice of $\rho(\omega)$ is a Gaussian mixture model; Since the number of modes of $\rho(\omega)$ is not a priori known, we will assume it to be infinite ($\rho(\omega) = \sum_{k=1}^{\infty} \pi_k \mathcal{N}(\omega|\mu_k,\Sigma_k)$ where $\sum_k\pi_k = 1$ and $\pi_k >0$), but given a finite dataset the model will realize only a finite number of Gaussians in the mixture.
We use a Dirichlet process (DP) prior on the components of the Gaussian mixture ($\pi$). 

The Dirichlet process is a distribution over discrete probability measures (i.e., atoms), $G=\sum_{k=1}^\infty \pi_k\delta_{\pi_k}$, with countably infinite support, where the finite-dimensional marginals are distributed according to a finite Dirichlet distribution \citep{Ferguson:1973uj}. 
We sample the mixture weights from a stick breaking prior,
i.e. $\pi \sim \mathit{GEM}(\alpha)$ where $\mathit{GEM}$ is the stick breaking prior \citep{sethuraman94}. We also put a Normal-Inverse-Wishart prior 
on the mean $\mu_k$ and variance $\Sigma_k$ of each of the Gaussian components. 

Secondly, model parameters are generated. In Section \ref{sec:randfeat} we discussed how functions in a kernel's RKHS can be approximated using a linear mapping in the random features. Thus, we consider models that operate linearly in the random features using a vector $\beta\in\R^{2M}$. As is standard in Bayesian regression and classification models \citep{Bishop2003}, we generate $\beta$ from a Normal prior, $\beta \sim \calN(\mu_{\beta},\sigma I)$.

Lastly, our observations are generated given a dataset $X := (x_1, \ldots , x_N)^T$ where each $x_i \in \R^d$. For example in regression tasks we have: 
\begin{equation}
y = g(x) + \epsilon\quad \epsilon \sim \calN(0,\sigma_{\epsilon}I),
\end{equation}
where $g$ is $\beta^T\varphi(x)$ and $\varphi(x)$ is calculated using \eqref{eq:rksfeats}. Thus $y \sim \calN(\beta^T\varphi(x),\sigma_{\epsilon})$. 



The complete generative model is given below and the corresponding plate diagram is shown in 
Figure \ref{fig:awesome_image}.

\begin{enumerate}
\item Draw the mixture weights $\pi$ over components of the kernel: $\pi \sim \mathit{GEM}(\alpha)$.
\item Draw the mixture components from Normal-Inverse-Wishart distribution. I.e. draw $\Sigma_k \sim \mathcal{W}^{-1}(\Psi_0,\nu_0),$ and $\mu_k \sim \mathcal{N}(\mu_0,\frac{1}{\kappa_0 } \Sigma_k)$ for $k=1,\ldots \infty$.
\item For each random frequency index $j=1,\ldots,M$
\begin{enumerate}
\item Draw the component from which the frequency vector is drawn. $Z_j \sim \mathit{Mult}(\pi)$.
\item Draw the corresponding random frequency vector $\W_j \sim \mathcal{N}(\omega|\mu_{Z_j}, \Sigma_{Z_j})$.
\end{enumerate}
\item Draw the weight vector, $\beta \sim \mathcal{N}(\mu_{\beta},\sigma I)$.
\item For each data point index $i=1,\ldots,N$
\begin{enumerate}
\item Define $\varphi(X_i)$ as in \eqref{eq:rksfeats}.
\item Draw the observation, e.g. for regression: $Y_i \sim \mathcal{N}(\varphi(X_i)^T \beta,\sigma_{\epsilon}I)$.
\end{enumerate}
\end{enumerate}

We note that the only change when going from regression to classification is in the step 5(b) of the generative procedure. This time we draw $Y_i$ from a sigmoid.

\begin{enumerate}
\item[5] For each data point index $i=1,\ldots,N$
\begin{enumerate}
\item[(b)] Draw the output binary label $Y_i \sim \sigma(\varphi(X_i)^T \beta)$, where $\sigma(x) = \frac{1}{1+\exp(-x)}$. 
\end{enumerate}
\end{enumerate}

\vspace{-0.15in}

\section{Inference}
\label{sec:inf}
\vspace{-0.05in}
We propose a MCMC based solution for inferring the parameters of the mixture of Gaussian distribution that defines $\rho(\omega)$. This includes finding the component assignment vector $Z$ and the mean and covariance $\mu_k$ and $\Sigma_k$ for each component. We will also sample the random frequencies $\W$ while marginalizing other parameters including $\pi$ and $\beta$ whenever possible. We will first describe the sampling equations for $Z,\ \mu_k,\ \Sigma_k$, which remain the same for both regression and classification. Afterwards we describe inference for $\W$ which depends on the specific application.


We want to sample from $p(Z,\mu,\Sigma,W | X,Y,\mathrm{rest})$, where $\mathrm{rest}$ are all the hyper-parameter of our model while other parameters including $\beta$ and $\pi$ have been integrated out. We use Gibbs sampling and sample each variable at a time given all other variables. 
\vspace{-0.05in}
\subsection{Sampling $Z_j$}
Recall that $Z_j$ indicates which component the random frequency $W_j$ is drawn from.
We use the Chinese restaurant process analogy to integrate out $\pi$, the component priors. Let $m_k \equiv \sum_l \delta(Z_l = k)$. The sampling equation for $Z_j$ can be derived from \citep{Neal:1998} and is shown below
\small
\begin{eqnarray}
 && P(Z_j = k | \mu,\Sigma,W,X,Y,\mathrm{rest}) = \\
  && \nonumber 
\left\{
\begin{array}{lr}
 \frac{m_k^{-j}}{M-1+\alpha}\mathcal{N}(\omega_j|\mu_k,\Sigma_k) & m_k^{-j}>0 \\
 \frac{\alpha}{M-1+\alpha}\int_{\mu,\Sigma}\mathcal{N}(\omega_j| \mu,\Sigma)\mathit{NIW}(\mu,\Sigma)d\mu d\Sigma & m_k^{-j} =0
\end{array}
\right. 
\end{eqnarray}
\normalsize
where $m_k^{-j} = \sum_{l: l\neq j} \delta(Z_l = k)$, $W_j = \omega_j$, $m_k^{-j}=0$ corresponds to unseen mixture component and $\mathit{NIW}$ is the Normal-Inverse-Wishart prior on mean and variance. 
\vspace{-0.05in}
\subsection{Sampling $\mu_k$ and $\Sigma_k$}\vspace{-0.05in}
Given the component assignment $Z$ and the random frequencies $W$, the posterior distribution of the covariance of each Gaussian component in the mixture is Inverse-Wishart, ie $\Sigma_k \sim \mathcal{W}^{-1}(\Psi_k,\nu_k)$ where $\Psi_k = \Psi_0 + \sum_{j: Z_j = k}^M (W_j - \overline{W}^{k})(W_j - \overline{W}^{k})^T + \frac{\kappa_0 m_k}{\kappa_0 + m_k}(\overline{W}^k - \mu_0)(\overline{W}^k - \mu_0)^T $, where $\overline{W}^k = \frac{1}{m_k} \sum_{j:Z_j=k}W_j$ and $\nu_k = \nu_0 + m_k$. Similarly, the posterior distribution of $\mu_k$ given, $\Sigma_k$, $Z$ and $W$ is a normal; i.e. $\mu_k \sim \mathcal{N}(\mu_k,\frac{1}{\kappa_k}\Sigma_k)$, where $\mu_k = \frac{\kappa_0\mu_0 + m_k \overline{W}^k}{\kappa_0 + m_k}$ and $\kappa_k = \kappa_0 + m_k$. See \citep{Gelman03} for details.
\vspace{-0.05in}
\subsection{Sampling $W$}\vspace{-0.05in}
We derive a Metropolis-Hasting (MH) sampler for sampling $W$. The posterior distribution of the random frequencies $W$ given the assignment $Z$, the parameters of the component $\mu$ and $\Sigma$, and the data, $X$ and $Y$ is proportional to 
\small
\begin{equation}
P(W|Z,\mu, \Sigma,Y,X,\mathrm{rest}) \propto P(W|Z,\mu,\Sigma)P(Y|X,W,\mathrm{rest}).
\label{eq:sampleW}
\end{equation} 
\normalsize
The first term in the LHS is a normal distribution $P(W|Z,\mu,\Sigma) = \prod_{j} \mathcal{N}(W_j | \mu_{Z_j}, \Sigma_{Z_j})$. Since it is difficult to sample directly from the posterior, we use MH, where the first factor of the RHS of \eqref{eq:sampleW} is used as a proposal distribution; i.e. $Q(W) = P(W|Z,\mu,\Sigma)$. Now, the acceptance ratio for a newly proposed $W^*$ is given by 
\small
\begin{align}
r = \min\left\{1,\frac{P(Y|X,W^*,\mathrm{rest})}{P(Y|X,W,\mathrm{rest})}\right\}. \label{eq:rreg}
\end{align}
\normalsize
Here the second term on RHS of \eqref{eq:rreg} is a ratio of model evidences and is calculated differently for regression and classification. 
\vspace{-0.05in}
\subsubsection{Regression}\vspace{-0.05in}
For regression we make use for conjugacy between prior of $\beta,\ \sigma_{\epsilon}$ and the likelihood to get a closed form solution for $P(Y|X,W,\mathrm{rest})$. In this case we sample $\sigma_{\epsilon}$ from $\mathrm{Inverse-Gamma}(a_0,b_0)$. The model evidence is then: 
\small
\begin{align}
 P(Y|X,W,\mathrm{rest}) 
&= \int_{\beta,\sigma_{\epsilon}}P(Y|W,X,\beta,\sigma_{\epsilon})P(\beta)P(\sigma_{\epsilon})d\beta d\sigma_{\epsilon} \nonumber \\
& \propto \frac{\Gamma(a_n)}{\Gamma(a_0)}\frac{b_0^{a_0}}{b_n^{a_n}}\sqrt{\frac{|\Lambda_0|}{|\Lambda_n|}} \label{ed:modev_int},
\end{align}
\vspace{-0.1cm}
\normalsize
where $\Lambda_0 = \frac{1}{\sigma^2}I$, $\varPhi(X) = (\varphi(X_1)^T \ldots \varphi(X_N)^T)^T$, $\Lambda_n = \varPhi(X)^T\varPhi(X) + \Lambda_0$, $\mu_n = \Lambda_n^{-1} (\Lambda_0\mu_{\beta} + \varPhi(X)^TY)$, $a_n = a_0 + \frac{n}{2}$ and $b_n = b_0 + \frac{1}{2}(Y^TY + \mu_0^T\Lambda_0\mu_0 - \mu_n^T\Lambda_n\mu_n)$. For more details refer to \citep{Minka99bayesianlinear}. 

It is worth noting that one may efficiently compute ratio's of model evidences if proposing a single $W_j$ at a time. That is, for each $j\in\{ 1, \ldots, M\}$ we propose $W_j^* \sim \mathcal{N}(W_j | \mu_{Z_j}, \Sigma_{Z_j})$ and calculate an acceptance ratio of 
\small
\begin{align}\vspace{-0.1in}
r_j = \min\left\{1,\frac{P(Y|X,W_j^*,W_{-j},\mathrm{rest})}{P(Y|X,W_j,W_{-j},\mathrm{rest})}\right\} \label{eq:rj}
\end{align}
\normalsize
where $W_{-j}= \{W_\ell\}_{\ell \neq j}$. This can be done efficiently because computing $P(Y|X,W_j^*,W_{-j},\mathrm{rest})$ only requires low-rank updates on $\varPhi(X)^T\varPhi(X)$, allowing for fast Cholesky updates.
\vspace{-0.05in}
\subsubsection{Classification}\vspace{-0.05in}
The aforementioned inference algorithm 
requires one to analytically obtain the model evidence of the data in terms of the model's random frequencies (eq: \ref{eq:rreg}, \ref{eq:rj}). However, a lack of conjugacy may make it intractable to marginalize other parameters to obtain the model evidence. For instance, the Gaussian prior on $\beta$ is not conjugate to a sigmoid. As a result it is difficult to directly compute the model evidence for a logistic regression model ie $P(Y|X,W,\mathrm{rest})$ where $\beta$ has been integrated out. Thus, for such situations where marginalization is intractable we must take a different approach to computing acceptance ratios for accepting random frequencies.

An approach one may take is to use a Laplace approximation to estimate the evidence as
\footnotesize
\begin{align}
&\log ( p(Y|X,W,\mathrm{rest})) \approx \log(p(Y|X,W,\bmap,\mathrm{rest})) \nonumber \\
&\qquad + \log(p(\bmap)) + \frac{N}{2}\log(2\pi) - \frac{1}{2}\log(|A|) \label{eq:lapclassaprrox} \\
& \where\ \bmap = \arg \min_{\beta} { \log ( P(Y|X,W,\beta, \mathrm{rest} ) P(\beta))} \nonumber \\
&\tand\ A = - \nabla^2 \log ( P(Y|X,W,\beta, \mathrm{rest} ) P(\beta))|_{\beta=\bmap}. \nonumber
\end{align}
\normalsize
However, there are a few drawbacks to using a Laplace approximation in this manner. First, due to approximation, one is no longer sampling from the true posterior. 
Second, calculating $\bmap$ when a closed form solution is not available (as with logistic regression) requires solving a costly $2M$ dimensional optimization problem when computing acceptance ratios.

In order to address these drawbacks whilst still mixing well we jointly sample the $j$th random frequency $W_j$ and the weight vector $\bc_j$ and $\bs_j$ corresponding to features $\cos(W_j^Tx)$ and $\sin(W_j^Tx)$ respectively. Specifically we sample from the joint distribution $W_j$ and $\beta_j^{\bullet} = \{ \bc, \bs \}$ given by:
\begin{eqnarray}
&& P(W_j,\beta_j^{\bullet}| Z,\mu, \Sigma,Y,X,\mathrm{rest})  \propto \label{eq:class_sample} \\ 
 && \quad P(W_j|Z,\mu, \Sigma) P(\beta_j^{\bullet }|\mathrm{rest}) P (Y|X,W_j,\beta_j^{\bullet}). \nonumber
\end{eqnarray}
\vspace{-0.1cm}
However, samples from \eqref{eq:class_sample} are not readily available, so we use Metropolis Hastings with the proposal distribution being
$$ Q = P(W_j|Z_j,\mu,\Sigma)\mathrm{Lap}(\beta_j^{\bullet} | X,Y, W_j,\mathrm{rest})$$
where $\mathrm{Lap}(\beta_j^{\bullet}|X,Y, W_j,\mathrm{rest})$ is the Laplace approximation of the posterior of $\beta_j^{\bullet}$, which requires only a 2-dimensional optimization:
\footnotesize
\begin{align}
\mathrm{Lap}(\beta_j^{\bullet}|X,Y, W_j,\mathrm{rest}) \approx P(\beta_j^{\bullet }|\mathrm{rest}) P (Y|X,W_j,\beta_j^{\bullet}). \label{eq:laplacej}
\end{align}  
\normalsize
Hence, acceptance ratio for jointly sampling a new $\{ W_j^*, \beta_j^{\bullet * } \}$ can be calculated as 
\footnotesize
$$\min \left\{ 1, \frac{P(Y|X,W_j^*,\beta_j^{\bullet*},rest)P(\beta_j^{\bullet*})\mathrm{Lap}(\beta_j^{\bullet} | X,Y,W_j)}{P(Y|X,W_j,\beta_j^{\bullet},rest)P(\beta_j^{\bullet})\mathrm{Lap}(\beta_j^{\bullet*}|X,Y,W_j^*)} \right\}.$$ 
\normalsize

\vspace{-0.15in}
\subsection{Runtime Complexity}\vspace{-0.05in}
We expound on the runtime complexity per iteration for the inference algorithms detailed above. Suppose that the $L$ is the number of components considered, $d$ is the data dimension, $M$ is the number of frequencies and $N$ is the number of data points. 
The runtime per iteration for sampling component parameters for both regression and classification is as follows: 1) sampling component parameters $\mu_\ell$`s and $\Sigma_\ell$`s (and maintaining stats): $O(L d^3)$; 2) sampling component assignments $Z_j$: $O(M L d^2)$.

For regression, sampling the random frequencies $W$ using low rank update takes $O( M (d^2 + d N + M N + M^2) )$. Thus, the total runtime per iteration is $O(M^2 d^2+M^2 N)= O(M^2 N)$ for large datasets where $N\gg M>d$, and $M \geq L$. 

For classification,
sampling $W_j$'s 
is $O\left( MN(d+\epsilon^-2) \right)$; where the $\epsilon^{-2}$ term arises from performing the 2-dimensional optimization required in \eqref{eq:laplacej} to $\epsilon$ precision \citep{cartis2010complexity}. Treating $\epsilon$ as a constant, we have a total runtime of $O(M N d)$ for inference.

Hence, we see that inference is linear in $N$ in either case, and so our method allows one to perform kernel learning in large datasets.



\vspace{-0.1in}\section{Experiments}\vspace{-0.1in}
We illustrate the use and performance of BaNK for both regression and classification on synthetic and real-world datasets below.
\vspace{-0.05in}
\subsection{Synthetic Data}
\vspace{-0.1in}

\begin{figure}[h]
\centering
\vspace{-0.1in}
    \subfigure[Synthetic data]{{\includegraphics[width=0.17\textwidth]{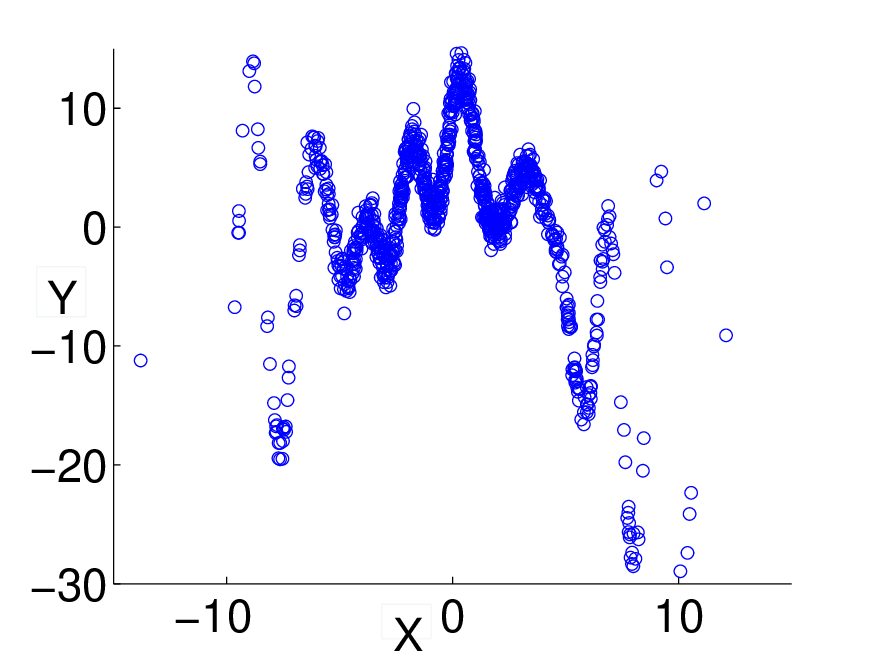}}}
    \subfigure[Kernel Estimate]{{\includegraphics[width=0.17\textwidth]{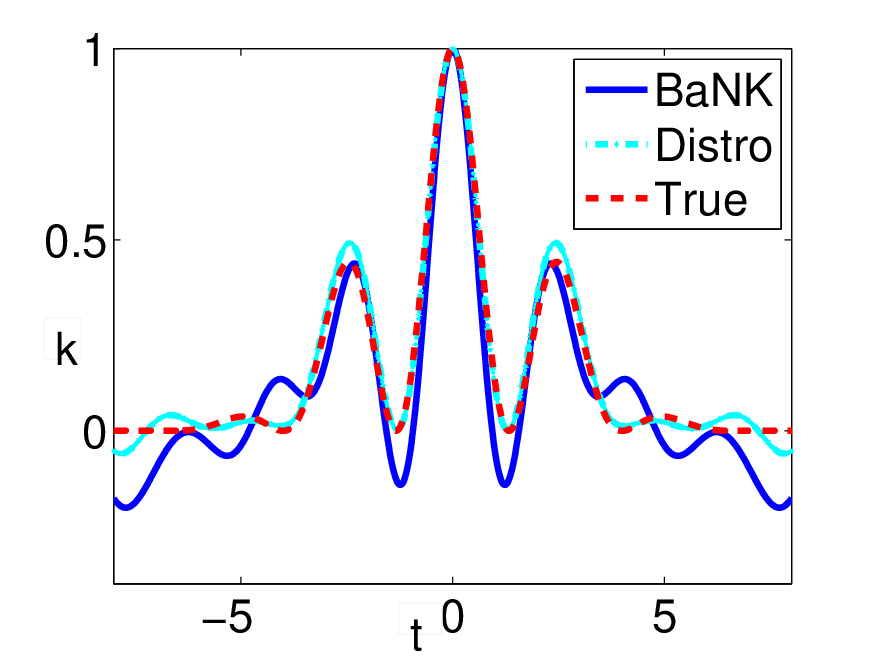}}\vspace{-0.1in}}\vspace{-0.1in}
    \caption{(a) Synthetic data used. (b) True $k$ in dashed red, $k$ estimated with true spectral distribution $\rho$ in cyan, and BaNK estimate in blue.} \label{fig:data1d}
\end{figure}

We give a simple 1-d kernel learning illustration with BaNK using synthetic data. We consider the shift-invariant kernel $k(t) = \exp\left(-\tfrac{1}{2(2^2)}t^2\right)(\tfrac{1}{2}+\tfrac{1}{2}\cos(\tfrac{3}{4}\pi t))$; that is, the kernel whose random frequency distribution is $\rho(\omega) = \tfrac{1}{2}\calN(\omega|0,\tfrac{1}{2^2}) + \tfrac{1}{2}\calN(\omega|\tfrac{3}{4}\pi,\tfrac{1}{2^2})$ (see Figure \ref{fig:data1d}). We look to learn the underlying kernel using 250 frequencies. We generated $N=1000$ instances $D=\{X_i,Y_i)\}_{i=1}^N$ where $X_i \iid \calN(0, 4^2)$, $Y_i \sim  \calN(\varphi_\rho(X_i)^T\beta, 1)$, with $\varphi_\rho$ being the random features from the kernel's true spectral distribution $\omega_j\iid\rho$ and $\beta \sim \calN(0,I)$. As explained above, using BaNK one may estimate $\rho$ by drawing from the posterior. We plot one such draw in Figure \ref{fig:data1d}(b). One can see that BaNK approximates the kernel rather well even though the underlying spectral distribution is multi-model, and the kernel is not easily decernable to the human eye based on the data plot (Figure \ref{fig:data1d}(a)).
\vspace{-0.1in}

\begin{table*}[ht]
\scriptsize
\centering
\begin{tabular}{{l}{c}{c}{c}{c}{c}{c}}
Dataset      & $N$ & $d$ & RKS & MKL & AlaC & BaNK \\
\hline
\verb|concrete|
& $1030$
& $8$
& $0.1313 \pm 0.0189$
& $0.0942 \pm 0.0100$
& $\pmb{ 0.0682 \pm 0.0092 }$*
& $0.1195 \pm 0.0108$
\\
\verb|noise|
& $1503$
& $5$
& $0.6974 \pm 0.0244$
& $\pmb{ 0.3217 \pm 0.0256 }$*
& $\pmb{0.3395 \pm 0.0489}$
& $\pmb{0.3359 \pm 0.0354}$
\\
\verb|prop|
& $11934$
& $16$
& $0.0006 \pm 3.6 \times 10^{-6}$
& $4.2 \times 10^{-5} \pm 5.7 \times 10^{-6}$
& $0.0003 \pm 0.0002$
&  $\pmb{ 8.9 \times 10^{-6} \pm 1.2 \times 10^{-6} }$*
\\
\verb|bike|
& $17379$
& $12$
& $0.1832 \pm 0.0049$
& $0.1509 \pm 0.0057$
& $\pmb{0.0467 \pm 0.0016}$*
&  $\pmb{0.0496 \pm 0.0022}$
\\
\verb|tom's|
& $28179$
& $96$
& $0.0479 \pm 0.0109$
& $0.0891 \pm 0.0100$
& $0.6583 \pm 0.0413$
&  $\pmb{0.0083 \pm  0.0018}$*
\\
\verb|cte|
& $53500$
& $386$
& $0.0886 \pm 0.0014$
& $0.0442 \pm 0.0003$
& $0.6223 \pm 0.0053$
&  $\pmb{0.0101 \pm  0.0003}$*
\\
\verb|music|
& $515345$
& $90$
& $0.8333 \pm 0.0028$
& $0.8524 \pm 0.0029$ 
& $0.7318 \pm 0.0705$
& $\pmb{0.7042 \pm 0.0038}$*
\\
\verb|twitter|
& $583250$
& $77$
& $0.3837 \pm 0.0397$
& $0.4572 \pm 0.0175$
& $0.2223 \pm 0.0358$
&  $\pmb{0.0981 \pm 0.0211}$*
\\
\end{tabular}
\captionof{table}{Regression MSE on UCI MLR. Asterisks denote the lowest MSE per dataset, methods in bold text were not found to be statistically different from the lowest MSE using a paired t-test with $p$-value $ < 0.05$. }\vspace{-.05in}
\label{tab:regres}
\end{table*}
\normalsize
\vspace{-.05in}

\begin{table*}[b!]
\scriptsize
\begin{center}

\label{tabl:classification_error}
\begin{tabular}{{l}{c}{c}{c}{c}{c}{c}}
Dataset      & $N$ & $d$ & RKS & MKL & RFO & BaNK \\
\hline
\verb|pima|  & $768$ & $8$ & $0.332 \pm 0.0201$  & $0.4455 \pm 0.0533$ & $\pmb{0.2592 \pm 0.0214}^*$ &$\pmb{0.263 \pm 0.0111}  $ \\ 
\verb|diabetic| & $1151$ &	$20$  & $ 0.3312 \pm 0.0083$ & $0.3051 \pm 0.0004 $  & $0.4263 \pm 0.0007$ & $\pmb{0.279 \pm 0.0022}^* $\\
\verb|eeg|  & $14980$ & $15$ & $0.0706 \pm 0.0019 $ & $\pmb{0.0544 \pm 0.0021}^*$ & $0.2411 \pm 0.1123 $ &$\pmb{0.0686 \pm 0.0012 }$\\
\verb|space| & $58000$ & $9$ & $0.0022 \pm 0.0004$ & $ \pmb{0.0015 \pm 0.0001}$ & $ 0.0018 \pm 0.00007$ & $ \pmb{0.0009 \pm 0.0001}^* $ \\
\verb|susy| & $100000$ & $18$ & $0.2009 \pm 0.0002 $ & $0.201 \pm 0.0001 $ &  $0.2089 \pm 0.00026$ & $\pmb{0.2005 \pm 0.0001 }^* $\\
\verb|skin| & $245053$ & $3$ & $0.1135 \pm 0.113$ & $ 0.2737 \pm 0.105$  & $ 0.0043 \pm 0.001$ & $ \pmb{0.0004 \pm 0.0001}^*$\\
\end{tabular}
\caption{Classification Prediction Error on UCI MLR. Asterisks denote the lowest error per dataset, while the methods in bold were not found to be statistically different from the lowest error using a McNemar's test \citep{NcNemar} with $p$-value $ < 0.05$. \label{tab:class_accuracy}}\vspace{-.05in}
\end{center}
\end{table*}
\normalsize
\vspace{-0.1in}
\subsection{Regression}
\label{sec:expreg}
\vspace{-0.05in}
Below we run experiments with various real-world datasets found in the UCI machine learning repository (UCI MLR)\footnote{\url{https://archive.ics.uci.edu/ml/index.html}}. We compare BaNK to a straight-forward random feature approach with a fixed kernel as well as other competitive random feature based kernel learning methods. In particular we compare to the following methods:

\textbf{RKS} For this method we take input covariates to be random features $\varphi(x_i)$ as in \eqref{eq:rksfeats}. Here we take the random frequencies $\omega_j \iid \calN(0,\sigma^{-2} I)$. This corresponds to approximating the RBF kernel: $K(x_i,x_l) = \exp\left(-\tfrac{1}{2\sigma^2}\norm{x_i-x_l}^2\right)$. Using these random features, we regress responses with ridge regression.

\textbf{MKL} One of the most widely used approaches to kernel-learning is multiple kernel learning (MKL) \citep{bach2004multiple,lanckriet2004learning}. Here, one attempts to learn a kernel using a non-negative linear combination of a fixed bank of kernels. That is, MKL attempts to learn a kernel $K$:
\footnotesize\vspace{-0.1in}
\begin{align}
K(x_i,x_l) = \sum_{m=1}^M \alpha_m K_m(x_i,x_l),\ \where\ \alpha_m \geq 0 \label{eq:MKL},
\end{align} 
\normalsize
and $K_1, \ldots, K_M$ are predefined kernels. The kernel weights $\alpha_m$ would then be optimized according to one's loss. Note that \eqref{eq:MKL} still requires the computation of a $N \times N$ Gram matrix, in fact, it requires $M$ such Gram matrices. However, we extend MKL to use random features and scale to larger datasets. If $K_m(x_i,x_l) \approx \varphi_m(x_i)^T \varphi_m(x_l)$, then 
\footnotesize\vspace{-0.1in}
\begin{gather}
K(x_i,x_l) \approx \sum_{m=1}^M \alpha_m \varphi_m(x_i)^T \varphi_m(x_l) = \bar{\varphi}(x_i)^T\bar{\varphi}(x_l),
\end{gather}
\normalsize
where $\bar{\varphi}(x_i) =  [\sqrt{\alpha_1} \varphi_1(x_i)^T, \ldots, \sqrt{\alpha_M} \varphi_M(x_i)^T]^T$. Hence, it is possible to work directly over input covariates of $\tilde{\varphi}(x_i) =  [\varphi_1(x_i)^T, \ldots, \varphi_M(x_i)^T]^T$, the concatenation of the random features for each kernel $K_1, \ldots, K_M$. We take our bank of kernels to be Laplace, RBF, and Cauchy kernels at various scalings. As with RKS, we regress responses through ridge regression.

\textbf{AlaC} Very recently, independent work by \cite{yang2015carte} has considered an optimization approach, called A la Carte, to learning a mixture of kernels. Here, an unconstrained, unpenalized, and non-convex GP likelihood problem is posed for regression and optimized over the parameters of a mixture model for random frequencies\footnote{Optimization was done using code provided by authors of \citep{yang2015carte}.}. 

We perform 5-fold cross-validation (picking parameters on validation sets and reporting back the error on test sets). For AlaC we cross-validate the total number of mixture components and frequencies per components for datasets with fewer than 100K instances; for larger datasets we use the suggested hyper-parameters in \citep{yang2015carte}. The total number of random features was chosen to be $768$ for RKS, MKL, and BaNK methods. For better interpretability, we standardized the output responses. In Table \ref{tab:regres} we report the mean squared error (MSE) $\pm$ standard errors. One may see that BaNK performs better or as well as other methods on nearly all the datasets. Furthermore, it seems like BaNK is better able to leverage larger datasets.
Lastly, we note that BaNK's sampling based approach with priors on mixture components seems more robust to local minima and over-fitting and has the ability to draw more frequencies from dominant components, which explains better performance w.r.t. Alac (e.g. for tom's dataset, Table \ref{tab:regres}). 



\subsection{Classification}
As previously mentioned, we may use the BaNK framework to perform kernel learning in classification tasks. Below we illustrate the use of BaNK for classification and kernel learning on real-world datasets from the UCI MLR. We compare the accuracies BaNK models achieve to traditional scalable kernel methods for classification; namely, we consider using the aforementioned RKS and MKL random features in a logistic model. 

Furthermore, we also compare to an optimization approach based on AlaC \citep{yang2015carte}, which we term random frequency optimization (RFO). 
Although it is possible to write and differentiate a data likelihood solely in terms of spectral density parameters (through the kernel they induce) for GP regression, a lack of conjugacy with a logistic likelihood and Gaussian priors requires approximate inference for classification. Thus, directly applying the approach of \cite{yang2015carte} will be troublesome.
To mitigate this difficulty, we jointly optimize a logistic loss both in terms of linear weights $\beta$ and spectral parameters $\{\nu, M, \mu\}$.  

Specifically, we minimize the following problem:
\tiny
\begin{align*}
& -\sum_{i=1}^N  
 Y_i \left\{ \sum_{k=1}^K \nu_k^2 \sum_{j=1}^D \left( \bc_{kj} \cos\left(\zeta_{ijk}\right) + \bs_{kj} \sin\left(\zeta_{ijk}\right) \right) + \beta_0\right\} \\
& +  \log\left[1 + \exp \left\{ \sum_{k=1}^K \nu_k^2 \sum_{j=1}^D \left( \bc_{kj} \cos\left(\zeta_{ijk}\right)  + \bs_{kj} \sin\left(\zeta_{ijk}\right) \right) + \beta_0 \right\}
\right] \\
&+ \frac{\lambda}{2} \left( \norm{\bc}^2 + \norm{\bs}^2 + \beta^2_0 \right),
\end{align*}
\normalsize
where $\zeta_{ijk} = X_i^T M_k w_{kj} + X_i^T\mu_k$ and $\bc\in\R^{KD}$, $\bs\in\R^{KD}$, $\beta_0\in\R$, $\nu\in\R^K$, $M_k\in\R^{d \times d}$, $\mu_k\in\R^d$ are optimized; $w_{kj} \in \R^d$ are standard Gaussian vectors that are drawn before optimizing and held fixed.

We again performed 5 fold cross validation and report the mean prediction error on test sets in Table \ref{tab:class_accuracy}. 
The results in Table \ref{tab:class_accuracy} show that BaNK consistently performed better or as well as the baselines.



\vspace{-0.05in}
\subsection{Timing Experiments}\vspace{-0.05in}

\begin{figure*}
\centering
    \subfigure[$N$ vs Error]{{\includegraphics[width=0.155\textwidth]{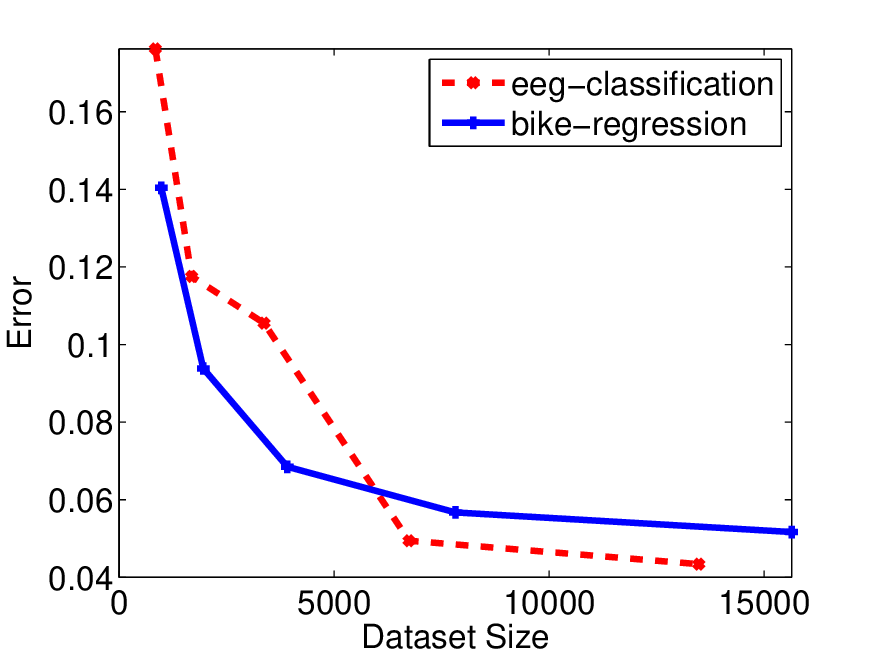}}\label{fig:nve}}
    \subfigure[$N$ vs Time ]{{\includegraphics[width=0.155\textwidth]{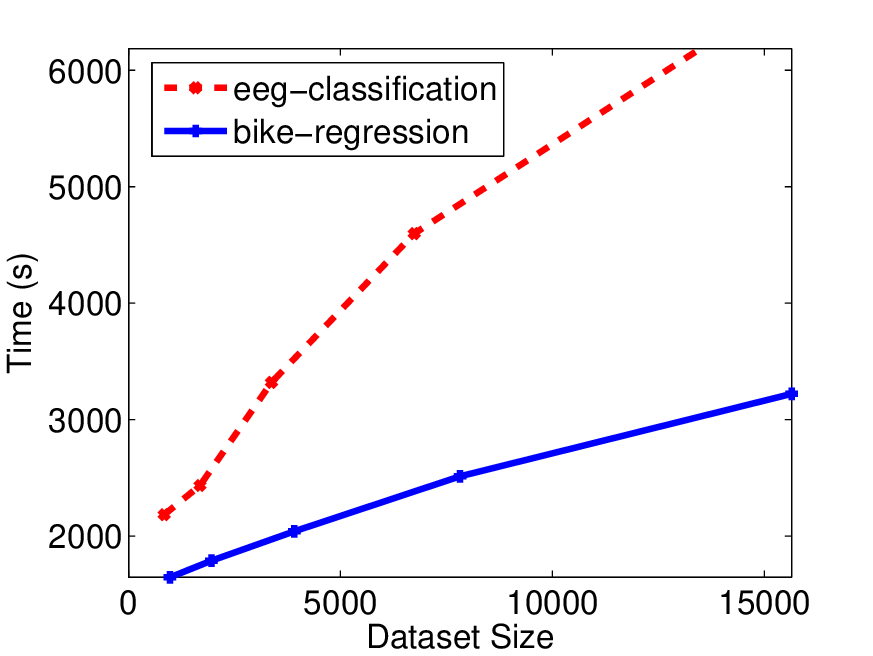}}\label{fig:nvt_s}}
    \subfigure[$N$ vs Time ]{{\includegraphics[width=0.155\textwidth]{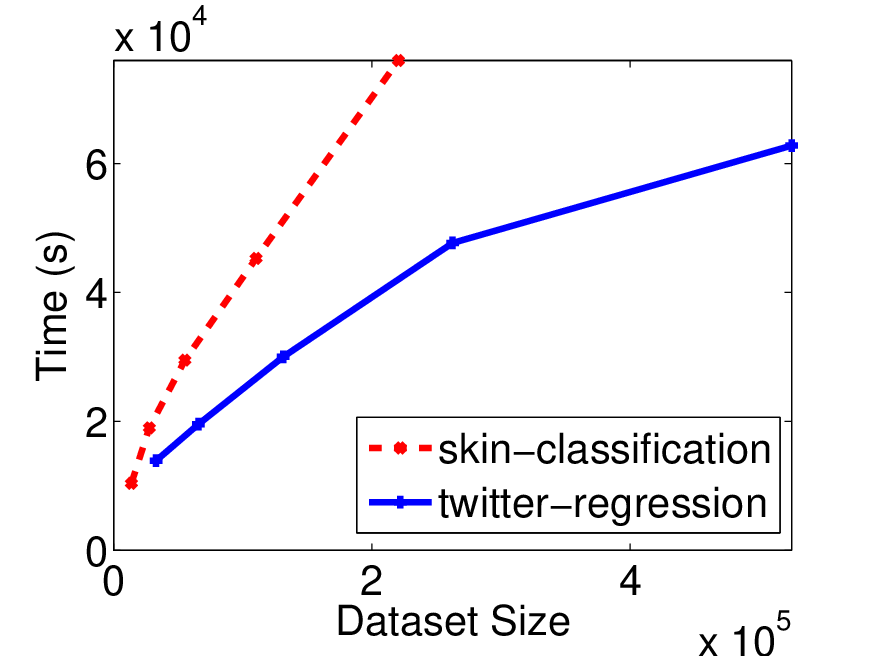}}\label{fig:nvt_l}}
    \subfigure[$M$ vs Error]{{\includegraphics[width=0.155\textwidth]{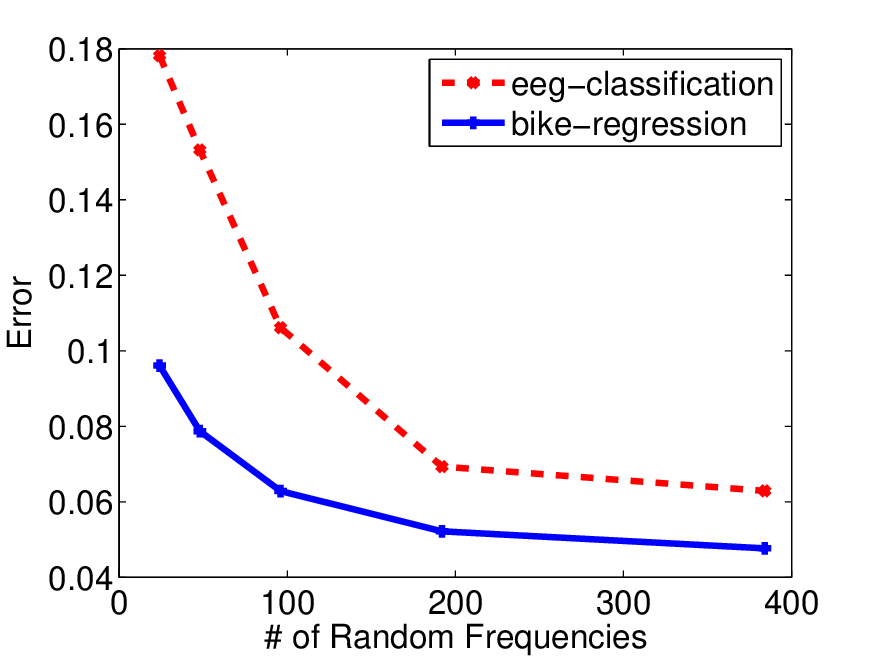}} \label{fig:mve}}
    \subfigure[$M$ vs Time ]{{\includegraphics[width=0.155\textwidth]{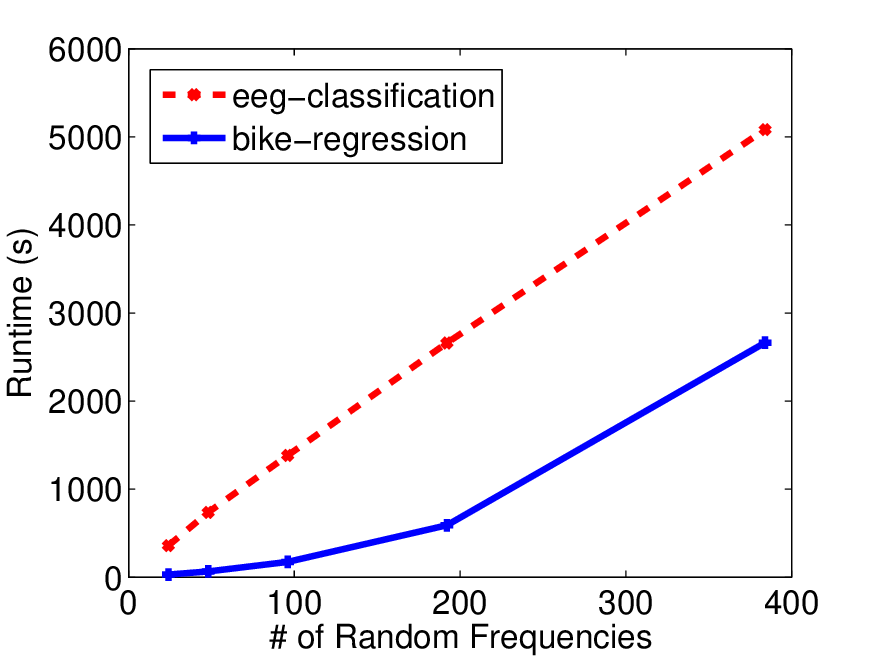}} \label{fig:mvt_s}}
    \subfigure[$M$ vs Time ]{{\includegraphics[width=0.155\textwidth]{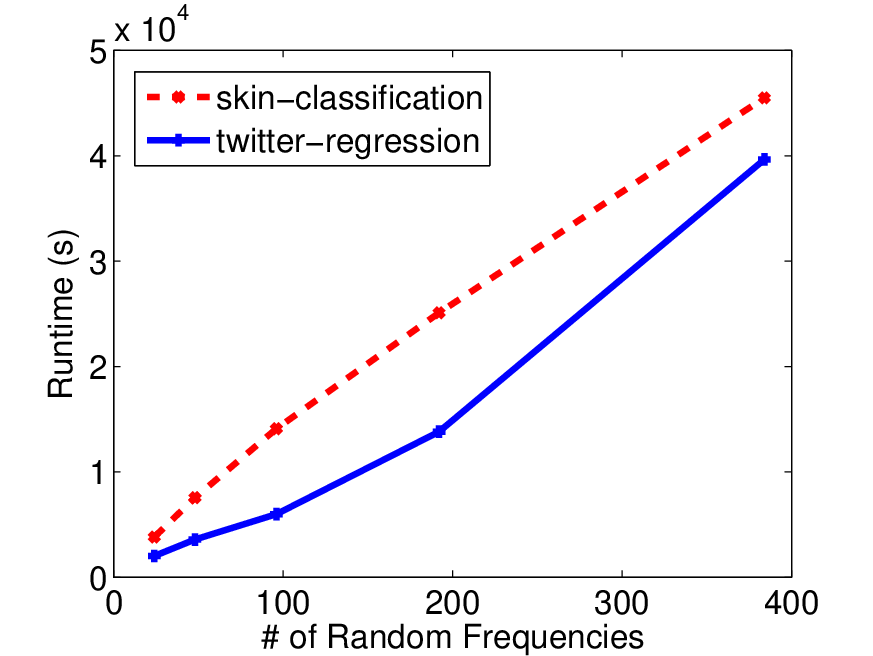}} \label{fig:mvt_l}}
    \caption{Runtime experiments. In all figures we denote classification curves with a dashed red line and regression curves with a solid blue line. Dataset names are shown in legends. Errors are prediction errors for classification tasks and MSE for regression tasks. Figure (a) shows how error changes with number of instances $N$; (b,c) shows the effect of increase in number of instances on runtime; (d) shows how error changes with number of random frequencies $M$; (e,f) show the effect of number of frequencies on computational time.  \label{fig:err_times}} \vspace{-0.1in}  
\end{figure*}

We empirically investigate the linear scaling of the BaNK method in terms of the number of instances $N$. It is this linear dependence that allows BaNK to perform kernel learning in large datasets, where naive kernel methods are generally $\Omega(N^2)$. We hold the number of random frequencies fixed at $M=384$ and vary $N$ to study the empirical dependence of runtime on dataset size for BaNK. We see the linear scaling is empirically verified in Figures \ref{fig:nvt_s} and \ref{fig:nvt_l}. Also, we observe lower test errors as dataset sizes increase indicating that BaNK is able to leverage more data to learn effective kernels and increase accuracy (Figure \ref{fig:nve}).

Furthermore, we illustrate the scaling of our BaNK method for regression and classification in terms of the number of random frequencies, $M$. As discussed in Section \ref{sec:inf}, the run-time complexity in large datasets will be $O(N M^2)$ for regression and $O(M N d)$ for classification.  We empirically study the dependence of $M$ on runtimes by varying $M \in \{ 24,\    48,\    96,\   192,\   384 \}$ and recording the runtime of BaNK using fixed datasets. Figures \ref{fig:mvt_l} and \ref{fig:mvt_s} show a quadratic growth in runtime for regression and a linear growth for classification. We see similar trends on other datasets. Moreover, we observe in Figure \ref{fig:mve} an increase in performance with diminishing returns as $M$ increases. Thus, we see that more frequencies aid kernel approximation and accuracy, but performance stabilizes after enough frequencies are chosen.

\begin{figure}
\centering
    \subfigure[Regression]{{\includegraphics[width=0.2\textwidth]{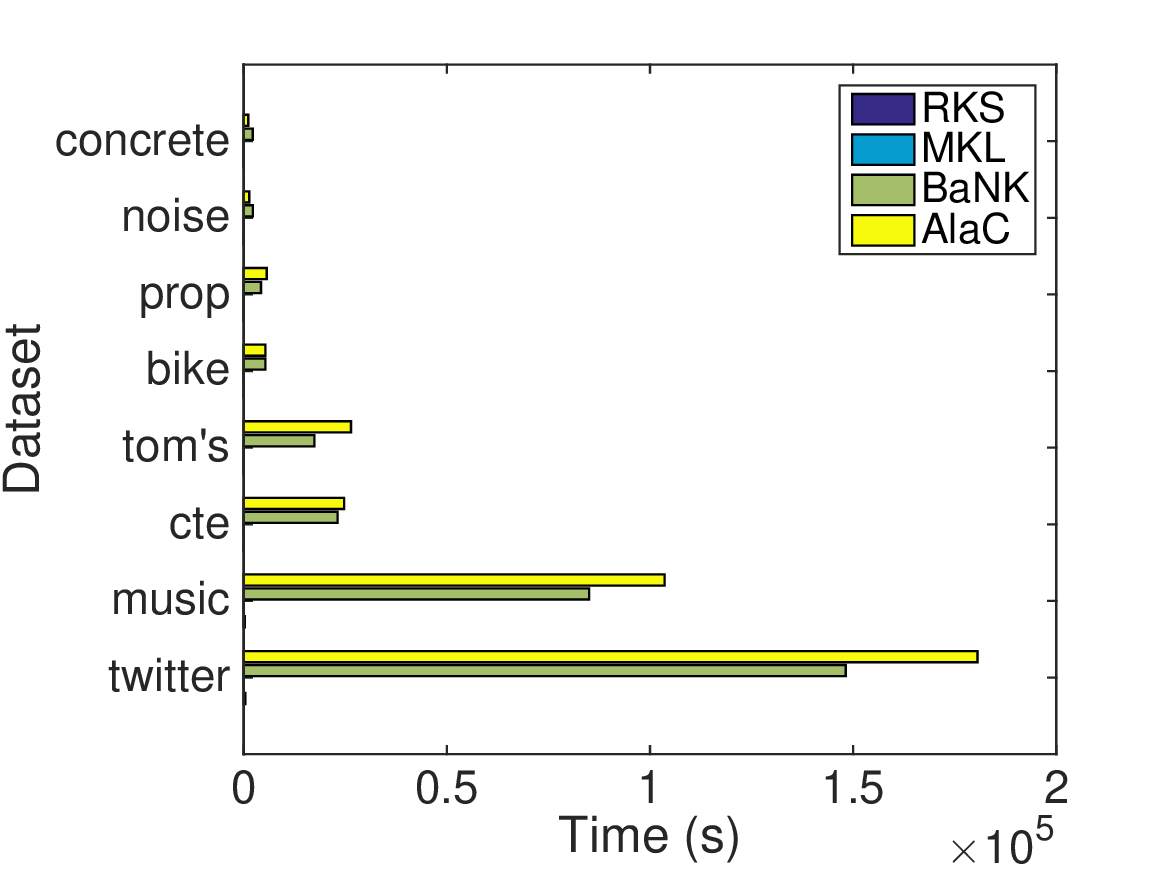}}}\vspace{-0.05in}  
    \subfigure[Classification]{{\includegraphics[width=0.2\textwidth]{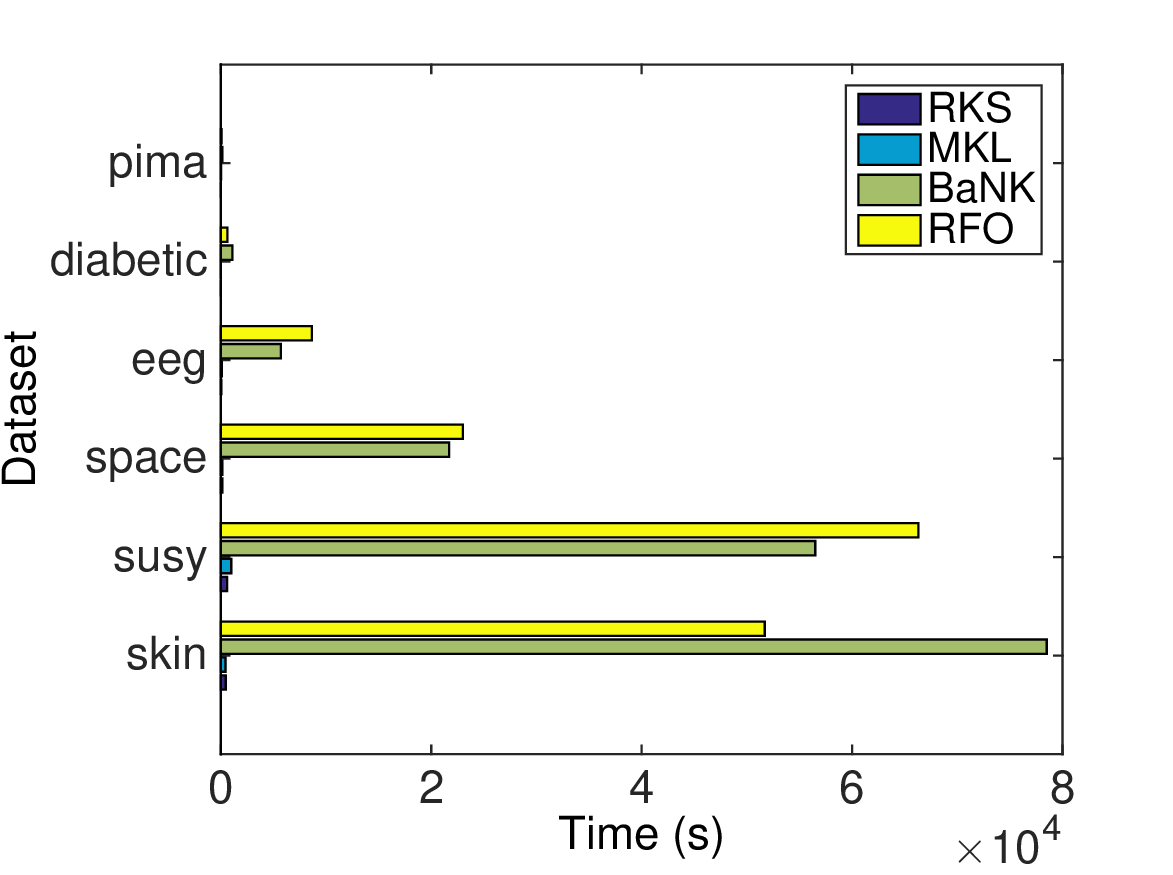}}}\vspace{-0.05in}  
    \caption{Runtimes for datasets. \label{fig:data_times}}\vspace{-0.1in}  
\end{figure}

Lastly, we record the runtimes on each dataset (Figure \ref{fig:data_times}) with the total number of random frequencies fixed at $M=384$ for all methods. 
While restricted methods that consider only a fixed set of random frequencies (RKS and MKL) perform fast, we see that BaNK's runtime is comparable to other methods that learn the random frequencies (AlaC and RFO) and can scale to large datasets. 

\vspace{-0.1in}  
\section{Related Work}
\vspace{-0.05in}
Given the poor scaling of kernel methods on datasets with many instances, several methods have looked at approximations of kernels that bypass the computation of large Gram matrices. For example, Nyst\"{o}m based methods \citep{drineas2005nystrom} look to give a fast to compute, low-rank approximation of the Gram matrix. Other approaches for summarizing the Gram matrix by the removal of elements or rows have been explored in \citep{scholkopf2002sampling, blum2006random, frieze2004fast, shen2006fast}. 
Furthermore, approximations based on KD-trees have been explored in \citep{shen2006fast}. 
In this paper, we work with kernel approximations based on random features, called Random Kitchen Sinks \citep{rahimi2007random, rahimi2009weighted, le2013fastfood}. As previously discussed, random feature approaches work using an empirical estimate of kernels that stem by drawing features from the Fourier transform of positive definite functions, which will be a distribution if properly scaled.

Kernel learning methods have also received attention due to the impact that kernel choice has on performance. Indeed, even if one fixes a family of kernels to use (e.g. RBF kernels) one still has to select the parameters of the kernels. This is often done with cross-validation or with methods like \citep{keerthi2007efficient,chapelle2002choosing}. A popular approach to learning kernels in a more flexible class is multiple kernel learning (MKL) \citep{bach2004multiple,lanckriet2004learning}. As illustrated is Section \ref{sec:expreg}, MKL learns a kernel that is the non-negative linear combination of a bank of fixed kernels.
However, naively applying MKL approaches would still require the computation of several large Gram matrices; instead, as previously discussed, one may combine random features to perform scalable MKL (see \citet{lu2014scale} for an application of such ideas). 
Very recently, independent work \citep{yang2015carte} has explored an optimization approach, A la carte, to learning parameters of mixture of kernels, where one optimizes the parameters generating random features in non-convex likelihood problems. See also \citep{buazuavan2012fourier} for another optimization approach.
Unfortunately, due to the non-convex nature of the optimization problem being optimized, such approaches yield non-interpretable kernels, whereas BaNK yield draws from a well defined posterior. 
\cite{Wilson2013} considers kernels as in \eqref{eq:mixkern}, but without inference over $L$, the number of mixture components. 
\cite{wilson2012process} mentions the possibility of putting a DP prior model on parameters, but does so without empirical details.

\vspace{-0.2in}
\section{Conclusion}
\vspace{-0.1in}
In this paper we propose an efficient and general data-driven framework, BaNK, for learning of kernels that scales to large datasets. By representing the spectral density using a non-parametric mixture of Gaussians, we capture a large class of kernels that can be learned. We provide a generative model for learning kernels while performing regression and classification tasks, and propose novel MCMC based sampling schemes to infer parameters of the mixtures. We show that our proposed framework outperforms other scalable kernel learning methods on a variety of real world datasets in both classification and regression task. 

\textbf{Acknowledgements}
This work is supported in part by NSF IIS-1247658,  NIH GWAS: R01GM087694, DOE DE-SC0011114 grants and an IBM Fellowship.


{\small 
        \bibliography{mainbib}}

\end{document}